\documentclass[sigconf,authorversion,screen]{acmart}
\usepackage{wrapfig}
\usepackage{enumitem}
\usepackage{multirow}
\AtBeginDocument{%
  \providecommand\BibTeX{{%
    \normalfont B\kern-0.5em{\scshape i\kern-0.25em b}\kern-0.8em\TeX}}}

\copyrightyear{2024}
\acmYear{2024}
\setcopyright{acmlicensed}\acmConference[UMAP '24]{Proceedings of the 32nd ACM Conference on User Modeling, Adaptation and Personalization}{July 1--4, 2024}{Cagliari, Italy}
\acmBooktitle{Proceedings of the 32nd ACM Conference on User Modeling, Adaptation and Personalization (UMAP '24), July 1--4, 2024, Cagliari, Italy}
\acmDOI{10.1145/3627043.3659553}
\acmISBN{979-8-4007-0433-8/24/07}

\begin{document}

\title{Negative Feedback for Music Personalization}

\author{M. Jeffrey Mei}
\email{jeffrey.mei@siriusxm.com}
\orcid{0000-0002-1083-598X}
\makeatletter
\@ifclasswith{acmart}{sigconf}{
\affiliation{%
  \institution{SiriusXM Radio Inc.}
  \streetaddress{1221 Avenue of the Americas}
  \city{New York}
  \state{New York}
  \country{USA}
  \postcode{10020}
}}{}
\makeatother

\author{Oliver Bembom}
\email{oliver.bembom@siriusxm.com}
\orcid{0009-0002-2617-5776}
\makeatletter
\@ifclasswith{acmart}{sigconf}{
\affiliation{%
  \institution{SiriusXM Radio Inc.}
  \streetaddress{1221 Avenue of the Americas}
  \city{New York}
  \state{New York}
  \country{USA}
  \postcode{10020}
}}{}
\makeatother

\author{Andreas F. Ehmann}
\email{andreas.ehmann@siriusxm.com}
\orcid{0009-0003-9589-0666}
\affiliation{%
  \institution{SiriusXM Radio Inc.}
  \streetaddress{1221 Avenue of the Americas}
  \city{New York}
  \state{New York}
  \country{USA}
  \postcode{10020}
}

% \renewcommand{\shortauthors}{Mei et al.}

\begin{abstract}
Next-item recommender systems are often trained using only positive feedback with randomly-sampled negative feedback. 
We show the benefits of using real negative feedback both as inputs into the user sequence and also as negative targets for training a next-song recommender system for internet radio.  
In particular, using explicit negative samples during training helps reduce training time by $\sim$60\% while also improving test accuracy by $6\%$; adding user skips as additional inputs also can considerably increase user coverage alongside improving accuracy. 
% Providing the user-selected target station allows the model to adapt its attention weights based on the target station, further improving performance by X\%. 
We test the impact of using a large number of random negative samples to capture a `harder' one and find that the test accuracy increases with more randomly-sampled negatives, but only to a point. Too many random negatives leads to false negatives that limits the lift, which is still lower than if using true negative feedback. 
We also find that the test accuracy is fairly robust with respect to the proportion of different feedback types, and compare the learned embeddings for different feedback types.
% many random negative samples would be required to approximate using hard negative samples and find this to be the order of $10^4$.
% We find that the learned embedding for implicit `skip' feedback is more similar to explicit negative feedback, but does share some similarity with explicit positive feedback.

% We evaluate the model over a range of dates following the training period and find the accuracy  decays at a rate of $\sim$1\% in 2 weeks, with faster decay when training without hard negatives. 
% Our work shows how explicit negative feedback can be used to improve both accuracy and coverage for music recommender systems.
\end{abstract}





% \received{1 June 2023}

\maketitle

\section{Introduction}

% Online music streaming services rely on recommender systems to filter a catalog with tens of millions of distinct tracks to those that match a listener's preferences. One way to account for these preferences, which may depend on a listener's context, is by offering stations that a listener can select, e.g. seeded by an artist or track. 
% This station context is important for music personalization as a listener may, for example, like both Rap and Country, but not both on the same station, and may lose interest if a recommender system repeatedly makes poor suggestions.
The rise of online streaming for music has led to millions, if not tens of millions, of distinct tracks being immediately accessible to users. 
However, users may only want to hear a small fraction of these tracks, so online music streaming services may rely on recommender systems to filter their catalogs for songs that are relevant to the user. 
Using each user's feedback to further personalize their song recommendations is crucial to increasing their engagement. 

This song recommendation can take the form of generating a static playlist, or a dynamic online radio station. In both cases, these song recommendations are provided in some sequence, but there are some salient differences between these sequences and other domains (e.g. e-commerce or natural language processing).
% User preferences may be station-specific - e.g., a listener may like both country and rap songs, but may not wish to hear country songs on a rap station and vice versa. 
% Although the set of songs played on a station (much like a playlist) are technically in a sequence, the feedback that is collected has some salient difference to sequences in other domains. 
Firstly, the sequential structure is much fuzzier - songs can be randomly swapped between positions to some extent (sometimes formalized in the user interface as `shuffle mode'), which weakens the positional effects. 
Secondly, certain feedback types are generally not repeatable, as a song that is explicitly liked (e.g. a `thumb-up') cannot be re-liked  without first being un-liked, and a song that is explicitly disliked (e.g. a `thumb-down') is generally not played again and cannot receive further feedback. 
This is not the case with all feedback types (e.g. songs can be skipped repeatedly, independently of prior feedback).% (and also independently of whether the song has been thumbed up in the past). 

For sequential modeling, most current research has coalesced around SASRec \cite{kang2018} or BERT4Rec \cite{sun2019}. Both methods use the Transformer architecture \cite{vaswani2017}.
SASRec trains using a causal filter where each $i$-th item in a sequence of $N$ total items is predicted using the preceding $i-1$ items, whereas BERT4Rec treats each training row as a cloze completion task for some $k$ positions, using the other $N-k$ items as inputs. 
SASRec uses a binary cross-entropy loss with a random item as the negative sample, which may lead to overconfidence in the output scores as a random negative is generally very easy to distinguish from the target item \cite{petrov2023}. 
BERT4Rec avoids this issue by calculating the softmax with the entire population as negatives, which comes at the cost of increased training complexity that may not be viable for industrial datasets that have millions of items in their catalog. 
There are ways around this, such as using sampled softmax as an estimate of the total loss \cite{bengio2008, wu2023}; custom loss functions that weight the random negatives differently \cite{petrov2023}; or trying to generate pseudo-negative samples, for example by choosing the top-$k$ negative samples as the hardest $k$ negatives for weight updates \cite{wilm2023} or more complex scoring methods to identify these harder negatives (e.g.\cite{weston2011, zhao2023}). 

Choosing a representative negative sample is thus a critical component of improving the training process for collaborative recommender systems. 
Often negative feedback is simply not collected; for example, in e-commerce it is easy to see what a user has clicked on and ordered, but it is not known if other items shown to the user were not clicked on because the user did not like them.  
However, if explicit negative feedback is collected, this data source can be a valuable source of negative sampling for training. For music recommendations, this negative feedback may be collected \textbf{explicitly} via a thumb-down button, or more \textbf{implicitly} like a skipped song, though skips may not always be a negative signal \cite{meggetto2023}.
Songs may receive both positive and negative feedback, such as if a user's taste changes or if a liked song is overplayed; hence, it is also important to treat the inputs as a sequence, with the caveat of certain feedback types being less sequential. 
These negative samples can be directly used in the training loss \cite{stoikov2021, wen2019}, or possibly as a complementary training task, whereby a model tries to minimize negative feedback instead of (or in addition to) maximizing positive feedback \cite{seshadri2023, wang2023}.

An issue with collaborative methods is the cold-start problem, both from an item perspective and a user perspective. Some items which are either new or unpopular do not have enough feedback to have a well-converged embedding; similarly, new or low-activity users do not have enough (or any) feedback to receive good recommendations via collaborative models. 
The former can be mitigated with additional data, such as partially sharing embeddings with similar items (e.g. songs by the same artist or items in the same e-commerce category) \cite{mei2023, wang2018} or by using content-based embeddings \cite{wang2014}. 
The latter is harder to overcome, but may be partially sidestepped by considering other data sources that may be more numerous, such as song skips or other contextual features like time/location \cite{dias2013, schedl2014}. 

Internet radio stations have some salient differences to other domains for next-item recommendation. 
Users' song preferences are highly contextual on the station: a user may like both jazz and rock songs, but not in the same session. 
Stations may therefore have some filtered set of eligible songs, which can be further ranked to personalize the station for the user. 
This also extends to user feedback: users may thumb up a song on one station and thumb down the same song on a different station. 
Because of this filtering, this means that songs on the same station are generally stylistically similar, and so it is harder to predict which songs from a given station a user may not like, as compared to other random songs that may be an entirely different genre or style.
The balance between maximizing songs that a user may like (true positives) versus minimizing songs that a user may dislike (false positives) is not well understood, and may be differ by user, or even differ for the same user in different contexts. These objectives may disagree with each other, such as in deciding whether to suggest more popular songs \cite{mena2020}. 
Nevertheless, the cost of bad recommendations may degrade the user experience (e.g. causing them to switch to a different provider). 

% One difficulty with such research is a lack of public datasets that include negative user feedback. For example, the 30Music \cite{30Music}, Last.FM \cite{celma2010} and KGRec-music \cite{oramas2016} datasets collect positive user feedback only.  The Piki dataset \cite{stoikov2021} is small (<$10^4$ users) but does contain explicit positive and negative user feedback.  
% The Spotify Sequential Skip Prediction dataset \cite{brost2019} contains plays and skips, which can be interpreted as positive and negative feedback respectively, but is at best an implicit signal with some notable differences to explicit negative feedback (e.g. \cite{meggetto2023}; also discussed in Section \ref{discuss:ftype}). It is also collected at the session, not user, level and has relatively short sequences (maximum length of 20). 
% Rating-based datasets in other domains, such as MovieLens are out of scope for this paper, as movie-watching habits may differ from music consumption habits (e.g. movies are rarely repeated, are not necessarily sequentially consumed, and are not necessarily consumed in thematic clusters like a radio station).

In this paper, we explore the benefits of incorporating additional feedback to a baseline transformer-based recommender system, both as inputs and as a source of negative samples. For brevity, we may refer to positive feedback (whether explicit or implicit) as `up' (`$+$') signals and explicit negative feedback as `down' (`$-$') signals, in addition to implicit `skip' (`$/$') signals.
% We find that skips can be incorporated as inputs which increases both the prediction accuracy of the test set and the user coverage. 
% Explicit negative feedback can also be included as hard negative samples, which both improves the test accuracy and reduces the training time.
% We find that shifting to a BERT4Rec-style cloze completion task allows for higher recommendation accuracy as the attention weights for the input songs can change considerably for different target stations.
The contributions of this paper are as follows:
\begin{itemize}
    \item We use explicit user-given hard negatives as negative samples to improve the test accuracy, and show that a hard negative can only be partially approximated by using the hardest negative from $k$ randomly-sampled negatives
    \item We show that using additional inputs (e.g. skips) improves personalization coverage while also improving accuracy, and explore the dependency of the prediction accuracy on the proportion of skips in a user's feedback
    \item We show the relationship between our test accuracy metric and metrics using true and false positives, and how these depend on the proportion of hard negatives used in training
    % \item We compare the learned feedback embeddings for positive feedback, negative feedback and skips and show that skipped songs are more similar, though still distinct, to explicit negative feedback 
    % \item We show that the recommendation accuracy decays with time at a rate of $\sim$1\% in 2 weeks, with faster decay if training without explicit hard negatives
\end{itemize}

\section{Data}
\ifdim\columnwidth<0.7\textwidth
\begin{table}[b]
\centering
\caption{\label{table:data}Summary statistics for the processed training sets. `$+$' denotes explicit positive feedback, `$-$' denotes explicit negative feedback and `$/$' denotes song skips. Statistics here are not necessarily representative of their user bases.}
\begin{tabular}{c|ccc}
\toprule
                       & Piki        &     Spotify      & Pandora     \\ \midrule
Granularity            & User      & Session      & User     \\
Feedback types         & $+$, $-$  & $+$, $/$     & $+$, $/$, $-$ \\
Ratio of $+$:$/$:$-$ & $1.1:0:1$    & $1:0.9:0$   & $4:13:1$    \\
Max. seq. length         & 400       & 20           & 400     \\
Median seq. length         & 24       & 9           & 289     \\
Max. lookback       &   3.7 years  & Same day   & 1 year    \\
Median lookback       &   2 days & Same day   & 9 months     \\
No. of sequences   & $8 \times 10^3$    & $8 \times 10^6$   & $10^7 $  \\
No. of tracks      & 2.5 $\times 10^5$ & 3 $\times 10^6$ & $10^6$ \\ \bottomrule
\end{tabular}
\end{table}
\else
\begin{wraptable}[16]{r}{0.5\textwidth}
\centering
\caption{\label{table:data}Summary statistics for the processed training sets. `$+$' denotes explicit positive feedback, `$-$' denotes explicit negative feedback and `$/$' denotes song skips. Statistics here are not necessarily representative of their user bases.}
\begin{tabular}{c|ccc}
\toprule
                       & Piki        &     Spotify      & Pandora     \\ \midrule
Granularity            & User      & Session      & User     \\
Feedback types         & $+$, $-$  & $+$, $/$     & $+$, $/$, $-$ \\
Ratio of $+$:$/$:$-$ & $1.1:0:1$    & $1:0.9:0$   & $4:13:1$    \\
Max. seq. length         & 400       & 20           & 400     \\
Median seq. length         & 24       & 9           & 289     \\
Max. lookback       &   3.7 years  & Same day   & 1 year    \\
Median lookback       &   2 days & Same day   & 9 months     \\
No. of sequences   & $8 \times 10^3$    & $8 \times 10^6$   & $10^7 $  \\
No. of tracks      & 2.5 $\times 10^5$ & 3 $\times 10^6$ & $10^6$ \\ \bottomrule
\end{tabular}
\end{wraptable}
\fi
We evaluate our model on three datasets. Two of them (Piki \cite{stoikov2021} and Spotify's Sequential Skip Prediction \cite{brost2019} datasets) are open-source and the third is a proprietary dataset from Pandora. 
Dataset statistics are summarized in Table \ref{table:data}. 
The Spotify dataset consists of plays and skips, which are not quite \textit{explicit} feedback, but can be mapped to positive and negative feedback respectively. It is a much larger dataset than the Piki one and is therefore included for comparison. 
The Piki dataset is filtered for only feedback coming from `personalized` recommendations. We assume that Piki playlists are comprised of similar songs (like a radio station); similarly, we filter the Spotify data to `radio'-type data only. 
This is necessary for our accuracy metric (Section \ref{sec:eval}), which is based off distinguishing which songs from a stylistically-consistent set of songs on some station may be liked by a user. 

% piki mean is 4 months; median is 2 days
% pandora mean is 9.5 months 

The Pandora dataset additionally has station information logged to provide context for each feedback, which is treated as another additive embedding \cite{mei2023}. The Piki and Spotify datasets, which do not have station information, are effectively trained with a null station embedding. 
% consists of user feedback (thumb up, skip, and thumb down) collected on each user's online radio stations over a one-year period ending 2023-07-31. These stations are themed based on a user selection of either a genre (e.g. `Classic Rock'), artist (e.g. `Beyonc\'{e}') or track (e.g. Adele's `Hello'). 
In all cases, the data is randomly split at the user level, with the final month of data time-separated for the test set, and 90\% of the remainder used for training and 10\% used for validation. 
The test sets consist of paired user feedback, consisting of one positive and one negative feedback (using explicit feedback where available, as for the Piki and Pandora datasets, and plays/skips in the Spotify dataset). 
For the Pandora dataset, the positive and negative  test samples are further required to come from the same station. 
We assume this condition is also satisfied by the Spotify dataset, which it is filtered for same-session `radio' tracks only. 
% Users in the training set are filtered for those with at least 5 up thumbs (and any number of skips), with the maximum number of inputs (total of skips and up thumbs) capped at 400. 
% This leaves the training set with $\sim$10M users, $\sim$1M songs, $\sim$1B up thumbs and $\sim$3B skips, with summary statistics given in Table \ref{table:stats}.

The Piki and Pandora datasets use explicit positive and negative feedback, with the Pandora dataset also having implicit negative feedback (skips). The Spotify dataset uses track plays and skips, which are more implicit, but can be thought of as positive and negative feedback. 
The Piki and Spotify datasets have a fairly even proportion of positive and negative feedback, whereas the Pandora feedback is dominated by skips. 
This is significant as although model \textit{training} requires users to have positive feedback (to use as training targets), there is no such requirement at \textit{inference}.

For all datasets, though less prominent in the Spotify dataset due to its smaller maximum input length, the distribution of user input sizes skews towards shorter sequences. 
This skew is stronger for the Piki dataset than the Pandora dataset and is reflected in the median sequence lengths in Table \ref{table:data}. 
To better align the training and test tasks, we enforce a condition that the final position will always be a prediction task (i.e. with no future feedback for inference). 
This means that training, validation and test sequences always end with an `up', to be used as a target for prediction.
Around one-third (32\%) of users in the Pandora dataset do not give any up thumbs at all, but do give some skips: these users are not included in the training set as they do not have positive examples for prediction, but can nevertheless be covered by a personalization model that accepts skips as inputs. A similar proportion (37\%) of users in the Spotify dataset also only give skips and are excluded from training.
% As some non-negligible proportion of users give no positive feedback but do use the `skip' button, 
This means that a model that accepts skips as inputs can increase the personalization rate, though the test accuracy for such skip-only users should be separately checked as they do not occur in the training set (Section \ref{discuss:ftype}). 
These high-skip users still contribute listener-hours and it is important to personalize their songs too.
% This means that a model that covers users with either skips or up thumbs will cover $32/68=47\%$  more users than a model that covers users with up thumbs only.

\section{Model Details}
The training architecture is a hybrid of SASRec and BERT4Rec: essentially, we combine the binary cross-entropy and general Transformer architecture of SASRec with the cloze task completion method of BERT4Rec. 
% For each position, the total embedding is the sum of song, station, position, and feedback (skip or thumb up) embeddings. 
For some proportion $p_{task}=0.15$ of `up' positions, we replace the `up' song with a mask token. 
Only the song attributes are masked; the target station at the prediction position is kept (null for Piki/Spotify) and the target feedback type is an `up'.
This reflects the real-life situation where we may seek to provide song recommendations for a user-selected station. 
The standard SASRec architecture cannot simulate this as it lacks mask tokens and the causal filter does not allow for any information at the prediction position to be used in training.

\ifdim\columnwidth<0.7\textwidth
\begin{figure}
  \includegraphics[width=0.43\textwidth]{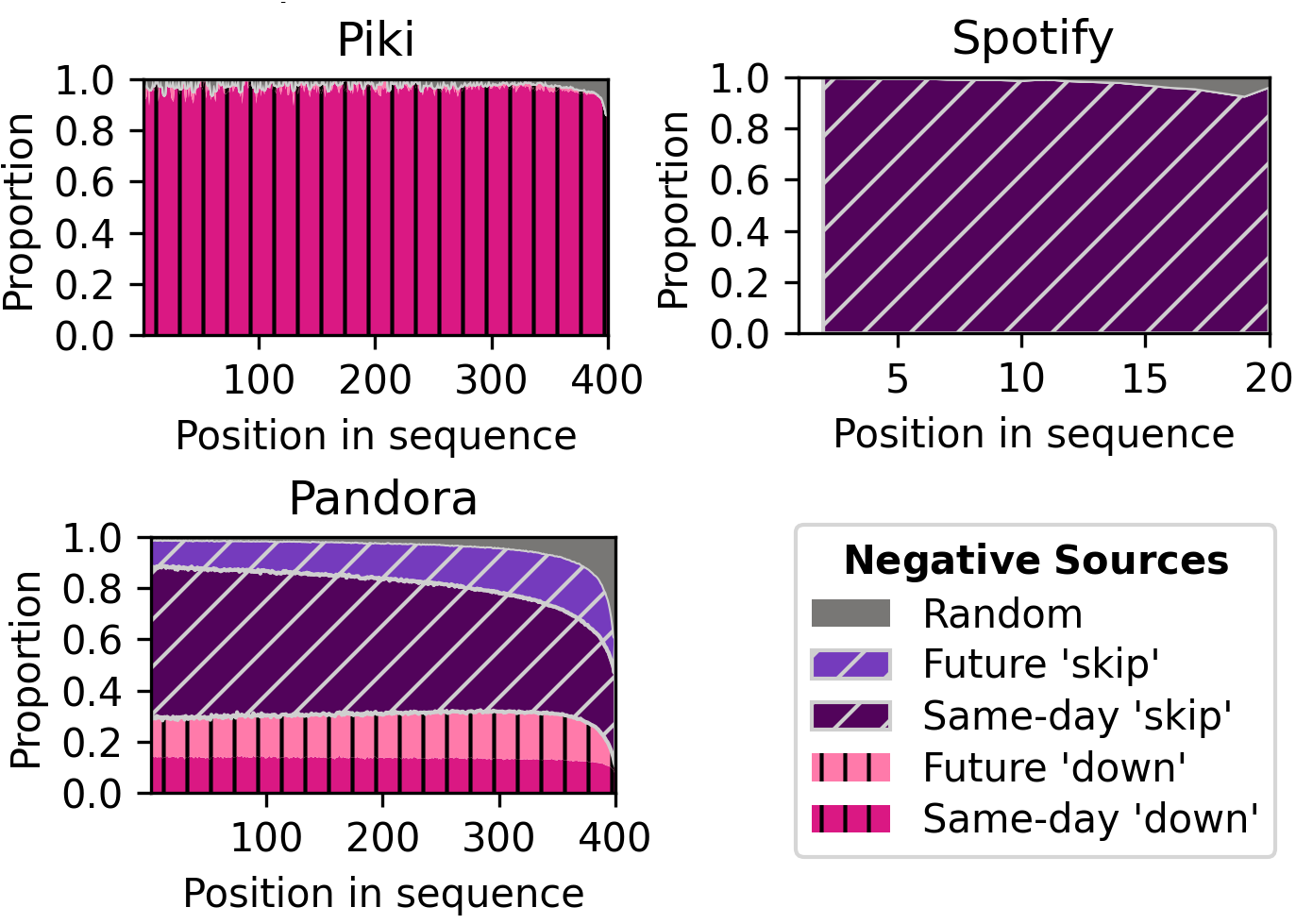}
  \caption{Proportion of sources for hard negative samples used at each position. More recent positions (higher $x$) are more likely to use randomly-selected negatives as there is less future feedback by definition.}
  \Description{A chart showing the distribution of sources for hard negative samples by position. Generally more skips are used than down thumbs, and more recent positions are more likely to lack future feedback and therefore use a random negative sample.}
  \label{fig:negs}
\end{figure}
\else
\begin{figure*}[b]
  \centering
  \includegraphics[width=0.90\textwidth]{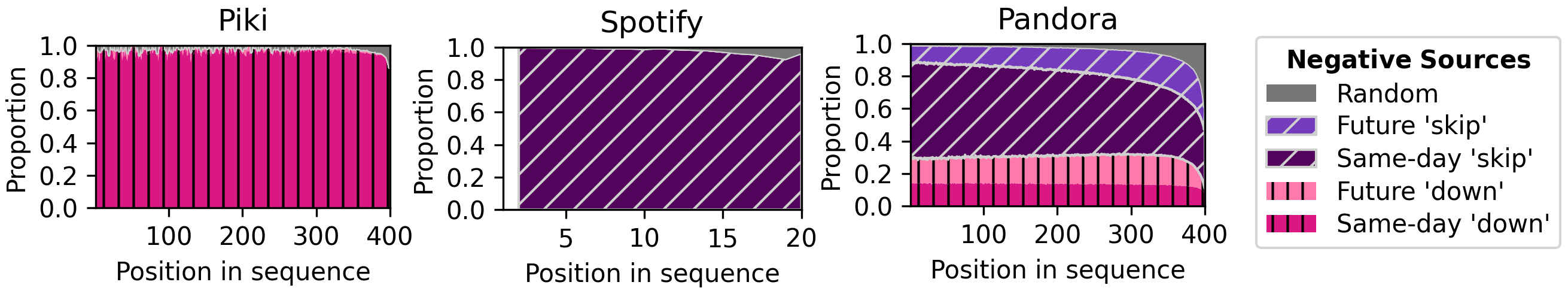}
  \caption{Proportion of sources for hard negative samples used at each position. More recent positions (higher $x$) are more likely to use randomly-selected negatives as there is less future feedback by definition.}
  \Description{A chart showing the distribution of sources for hard negative samples by position. Generally more skips are used than down thumbs, and more recent positions are more likely to lack future feedback and therefore use a random negative sample.}
  \label{fig:negs}
\end{figure*}
\fi
For the hard negative sample at each masked position, we use a \textbf{same-day down thumb} from the same station (if one exists), otherwise a\textbf{ same-day }\textbf{skip} from the same station (if it exists), otherwise any \textbf{future} \textbf{down} thumb (if it exists), otherwise any \textbf{future} \textbf{skip} (if it exists), otherwise a \textbf{random} song (Fig. \ref{fig:negs}). 
For the Spotify dataset, this simplifies greatly as there are no down thumbs and only same-day feedback (Fig. \ref{fig:negs}). 
Similarly, the Piki dataset has no skips and has a median sequence length of 2 days (Table \ref{table:data}) so rarely has future feedback to use. 
For all datasets, future feedback is less frequent for more recent positions, which are therefore more likely to use random negative samples  (Fig. \ref{fig:negs}).
The proportion of hard negatives used during training can be further controlled with a parameter $p_{hard}$, which is the proportion of training positions that use a hard negative. 
We discuss the impact of changing $p_{hard}$ in Section \ref{sec:hardneg}.
`Skips' and `downs' that are used as negative samples are removed from the user's input sequence.
To distinguish feedback types in the input, we use a learned additive embedding mapped to the different feedback types (discussed further in Section \ref{discuss:ftype}). The total embedding at each position is therefore the sum of the song, station (null for Piki and Spotify datasets), feedback and positional embeddings.

The \textbf{baseline} model uses only positive feedback as inputs (i.e. no feedback embedding) and uses $p_{hard}=0$ (randomly-sampled instead of user-provided hard negatives during training).
Generally, these random samples are likely to be from an unrelated genre and thus are too easy for the model to rank (leading to the overconfidence issue identified by Petrov \& Macdonald (2023) \cite{petrov2023}). 
In the absence of real negative user feedback, one can attempt to sample more negatives (up to the entire set of $N$ items, like BERT4Rec) to try and capture some hard(er) negative(s); this is explored in Section \ref{sec:hardneg}.

\section{Evaluation}\label{sec:eval}
The model is evaluated against real user data from the subsequent month after the training period, using users who have given both positive and negative feedback (on the same station for Pandora). 
Because songs (whether thumbed up or down) that are played on the same station are fairly similar in musical style, this is a much harder problem than simply comparing an `up' song against a random negative sample. 
Generally, even `down' songs by some user are thumbed up by many other users, making them harder to distinguish than a random negative sample.

An accurate prediction is when the `up' song is scored higher than the `down' song; a model that randomly guesses would therefore have an accuracy of $0.5$.  
Our `test accuracy' is therefore equivalent to the probability that a random pair of positive/negative feedback is correctly ranked, which is equivalent to the average area, of all users, under the receiver operating characteristic curve (\textbf{AUC}) \cite{hanley1982}. 
% For this harder task, the baseline model has an accuracy of 0.68. 
If testing against random negatives, both our model and the baseline have an accuracy $\gg 0.99$. 
This is not very useful, and also not very relevant to the actual usage of these models, which is to distinguish and rank songs that are on a given station (i.e. songs that are generally similar) for each user. 
The link between test accuracy and other metrics that focus on true positives (e.g. recall) is explored further in Section \ref{discuss:mrr}.

\section{Results and Discussion}
For the Spotify and Pandora datasets, we find that incorporating additional feedback (both as inputs and as hard negative samples during training) improves the test set accuracy by \textbf{$6\%$} as compared to a model that is only trained on positive feedback and random negatives. 
The Piki dataset has a similar lift but due to its smaller size, the lift is not statistically significant.
By far, the biggest contribution to the lift comes from incorporating hard negatives into the loss function. 
We explore the impact of hard negatives further in Section \ref{sec:hardneg}, and the impact of additional feedback types in \ref{discuss:ftype}.
% We find the highest test accuracy for $p_{hard}=0.2$, although the accuracy is fairly robust with respect to $p_{hard}$ (discussed more extensively in Section \ref{sec:hardneg}). 
% Other metrics like the mean reciprocal rank are not as relevant for our use case unless we also consider the ranks of the hard negatives, as discussed further in Section \ref{discuss:mrr}.

% \begin{figure}[h!]
%   \includegraphics[width=0.5\textwidth]{figs/decay_overall.png}
%   \caption{Accuracy deteriorates at a rate of approximately 1\% per 2 weeks. There is a weekly periodicity to thumb rates, and so the decay should be calculated using the weekly peaks.}
%   \Description{blah}
%   \label{fig:decay}
% \end{figure}

\ifdim\columnwidth<0.7\textwidth

\begin{table}
\centering
\caption{\label{table:ablation}Ablation study with percentage point changes in accuracy. 
The Piki data is too small for any significant results and is omitted.
%The biggest contribution is from using hard negative samples.
}
\begin{tabular}{@{}cc@{}cc}
\toprule
 \multirow{2}{*}{Method}                 & \multicolumn{2}{c}{$\Delta$ Accuracy} \\ 
                      &  Spotify   &\ \ Pandora    \\ \midrule 
No positional emb.     & $-0.6\%$    &  $-0.7\%  $ \\
No hard negatives      & $-5.4\%  $  & $ -6.3\% $  \\
Only positive inputs    & $-0.6\% $  &  $-0.5\% $  \\
Half max. seq. length   & $+0.1\%$   & $ -0.2\%$   \\ \bottomrule
\end{tabular}
\end{table}
\else
\begin{wraptable}[10]{r}{0.402\textwidth}
\centering
\caption{\label{table:ablation}Ablation study with percentage point changes in accuracy.
The Piki data is too small for any significant results and is omitted.
%The biggest contribution is from using hard negative samples.
}
\begin{tabular}{@{}cc@{}cc}
\toprule
 \multirow{2}{*}{Method}                 & \multicolumn{2}{c}{$\Delta$ Accuracy} \\ 
                      &  Spotify   &\ \ Pandora    \\ \midrule 
No positional emb.     &$ -0.6\%  $  &  $-0.7\% $  \\
No hard negatives      &$ -5.4\% $   &  $-6.3\% $  \\
Only positive inputs    & $-0.6\%$   &  $-0.5\%$   \\
Half max. seq. length   &$ +0.1\%$   &  $-0.2\% $  \\ \bottomrule
\end{tabular}
\end{wraptable}
\fi

\subsection{Ablation study}\label{sec:ablation}
The overall lift in accuracy after incorporating negative feedback ($6\%$) predominantly comes from the use of hard negatives in training (Table \ref{table:ablation}).  
The lift from adding additional feedback types into the user inputs, although positive, is an order of magnitude smaller. 
An additional benefit of including skips as inputs is in increasing the coverage, by $50\%$ for Pandora and $37\%$ for Spotify, of the model to skip-only users.
The benefit of positional embeddings is similar in magnitude for both datasets. 
Halving the maximum sequence length slightly improves the Spotify accuracy while slightly reducing the Pandora accuracy; this may arise from differences between noisier implicit (`play') and explicit (`up') positive feedback.
% Halving the maximum sequence length has an even smaller impact, particularly for the Spotify data, as most sequences have lengths far below the maximum input length. 
% In Sections \ref{sec:hardneg}-\ref{discuss:pos} we look into these impacts in greater detail.
% As our model features many additional enhancements to tehe base SASRec Transformer model, we conduct an ablation study to quantify the impact of each individual part. We find that the biggest contributions to the lift in accuracy come from using hard negatives, using positional embeddings and using target station embeddings. These are discussed in turn in Sections \ref{sec:hardneg}-\ref{discuss:seed}.

\subsection{Hard negative samples}\label{sec:hardneg}
Generally, a hard negative will have a higher score (and thus be harder to distinguish from the actual positive sample) than an unrelated random negative sample. 
It is not necessarily the highest-scoring negative sample possible, as other songs (e.g. by the same artist as the positive sample that are not yet known to the user) may have higher scores, although we do not necessarily know if the user would like or dislike those songs. 
This means that generally, for $k \ll N$, $k$ random-chosen negatives are likely to all have lower scores than a hard negative. 
Conversely, this means that we may estimate the value of a hard negative by comparing the performance of a model trained with \textbf{one} hard negative per positive sample to one that is trained with \textbf{$k$} randomly-selected negatives and using the top-scoring (i.e. hardest) one for computing the loss and comparing the test accuracy for different $k$.

\ifdim\columnwidth<0.7\textwidth
\begin{figure}[h]
 \includegraphics[width=0.33\textwidth]{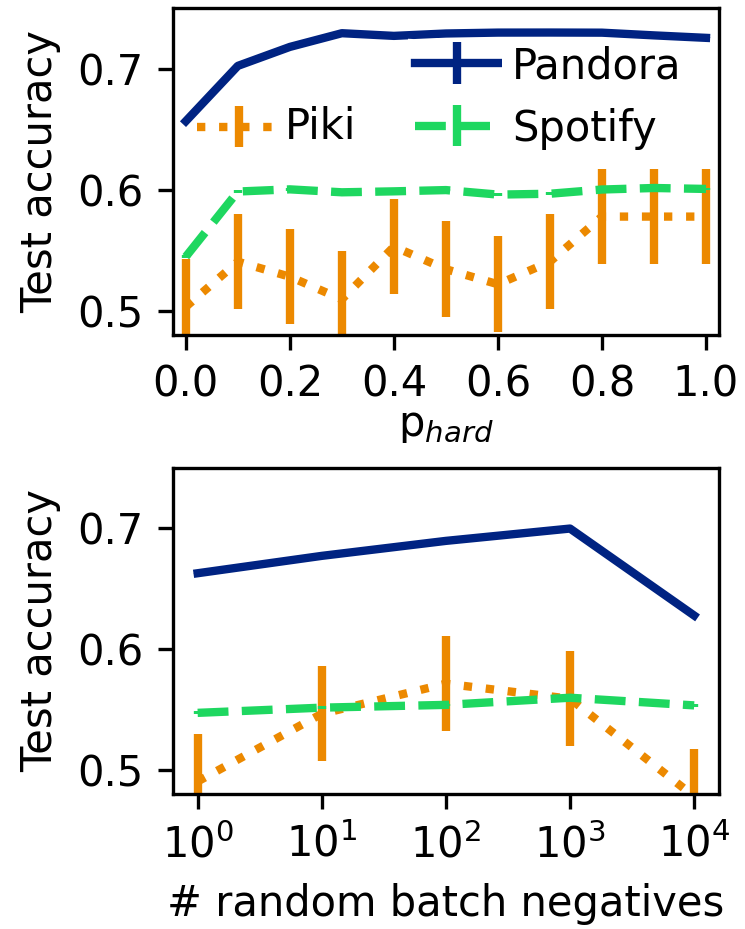}
  \Description{
  Top: a chart showing the general increase in test accuracy as the proportion of hard negatives increases. 
  Bottom: a chart showing the increase with accuracy as more random negative samples are used up to 10$^3$ (10$^2$ for Piki), and then a decline for higher values. 
  % Bottom: A chart showing a big jump in accuracy for non-zero $p_{hard}$, and also generally much faster training times when using higher $p_{hard}$.
  }
\caption{Top: The test accuracy generally increases with higher $p_{hard}$. 
% The Spotify and Pandora datasets reach a plateau in test accuracy after $p_{hard} = 0.2$ and $0.3$ respectively. The Piki dataset is much noisier but generally has higher test accuracy for $p_{hard} \gg 0$.
  Bottom: Test accuracy increases with more batched random negatives, but using too many hurts performance. The maximum lift is also smaller than using a true hard negative sample.}
  \label{fig:hard_negs}
  \end{figure}
\else
\begin{wrapfigure}[11]{R}{0.5\textwidth}
  \includegraphics[width=0.48\textwidth]{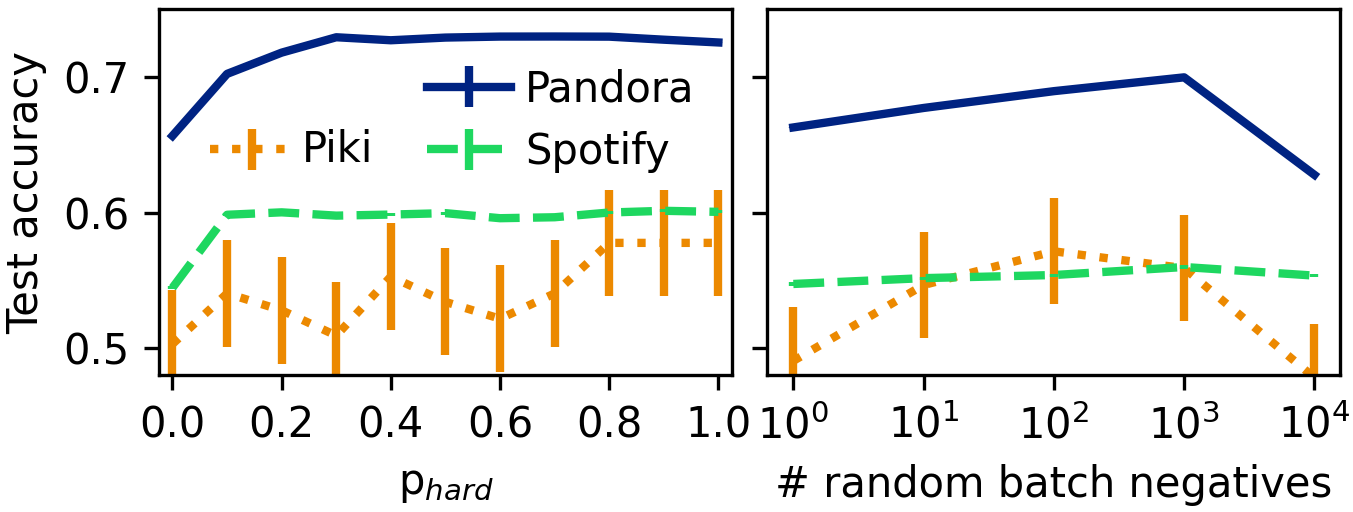}
  \Description{Left: a chart showing the general increase in test accuracy as the proportion of hard negatives increases. 
  Right: a chart showing the increase with accuracy as more random negative samples are used up to 10$^3$ (10$^2$ for Piki), and then a decline for higher values. }
\caption{Left: The test accuracy generally increases with higher $p_{hard}$. 
% The Spotify and Pandora datasets reach a plateau in test accuracy after $p_{hard} = 0.2$ and $0.3$ respectively. The Piki dataset is much noisier but generally has higher test accuracy for $p_{hard} \gg 0$.
  Right: Test accuracy increases with more batched random negatives, but using too many hurts performance. The maximum lift is also smaller than using a true hard negative sample.}
  % In the absence of a hard negative, we find using multiple random negative samples and taking the one with the highest loss can increase the test accuracy, but not enough to match a model trained with hard negatives. Too many random negative samples may lead to false negatives being used in training and reduce accuracy.}
  \label{fig:hard_negs}
\end{wrapfigure}
\fi

\subsubsection{Test accuracy}
We find that the test accuracy increases with higher $k$, but peaks around $k=10^3$ for the Pandora and Spotify datasets and $k=10^2$ for the Piki dataset, and then drops beyond that (Fig. \ref{fig:hard_negs}). 
This suggests that with too many random negatives, the highest-scoring negative is potentially a false negative, which worsens model performance. 
The Piki dataset has fewer songs by an order of magnitude, which may be why the $k$ turning point is also an order of magnitude lower than the Spotify/Pandora datasets.
Important to note is that the best test accuracy using batched random negatives is still lower than the best test accuracy using explicit hard negative samples. 
Further work can be done to implement other ways of distinguishing true and false negatives \cite[e.g.][]{weston2011, zhao2023}.
% Although there are other ways to attempt to distinguish true and false negatives, they still have some associated error and are no substitute for explicit user signals. 
% In general, our results suggest that methods that increase the difficulty of sampled negatives can improve test accuracy, but worse than using a true hard negative. Too many random negative samples may also introduce false negatives that reduce model accuracy. 
For datasets that lack explicit negatives, more research should be done on how to determine if/how sampling too many negatives can hurt performance.
% with $\sim$10$^4$ random negatives needing to be sampled before finding a similarly-hard negative that allows for a similar test accuracy. 

\ifdim\columnwidth<0.7\textwidth
\begin{figure}[t]
  \includegraphics[width=0.33\textwidth]{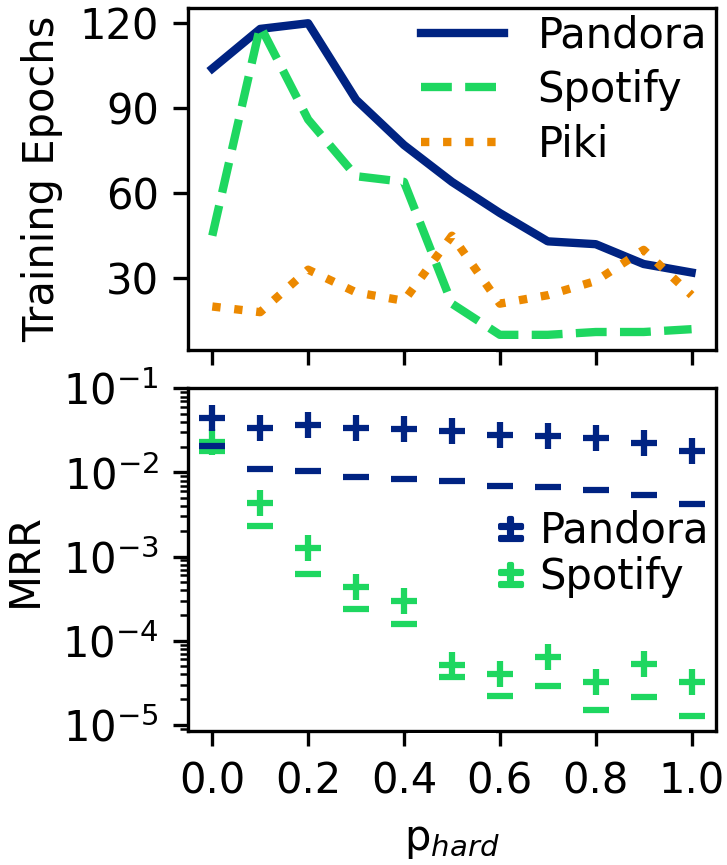}
  \caption{For higher $p_{hard}$, training time tends to decrease (top), while the mean reciprocal ranks of the positive ($+$) and negative ($-$) samples generally both decrease (bottom), even as the test accuracy increases (Fig. \ref{fig:hard_negs}). The Piki dataset is too small/noisy to analyze and is omitted.}
  \Description{Top: A chart showing how training time decreases when using higher $p_{hard}$. Bottom: A chart showing how the ranks of both the positive and negative songs decline as $p_{hard}$ increases.}
  \label{fig:epochs}
\end{figure}
\else
\begin{wrapfigure}[17]{r}{0.34\textwidth}
  \centering
  \includegraphics[width=0.3\textwidth]{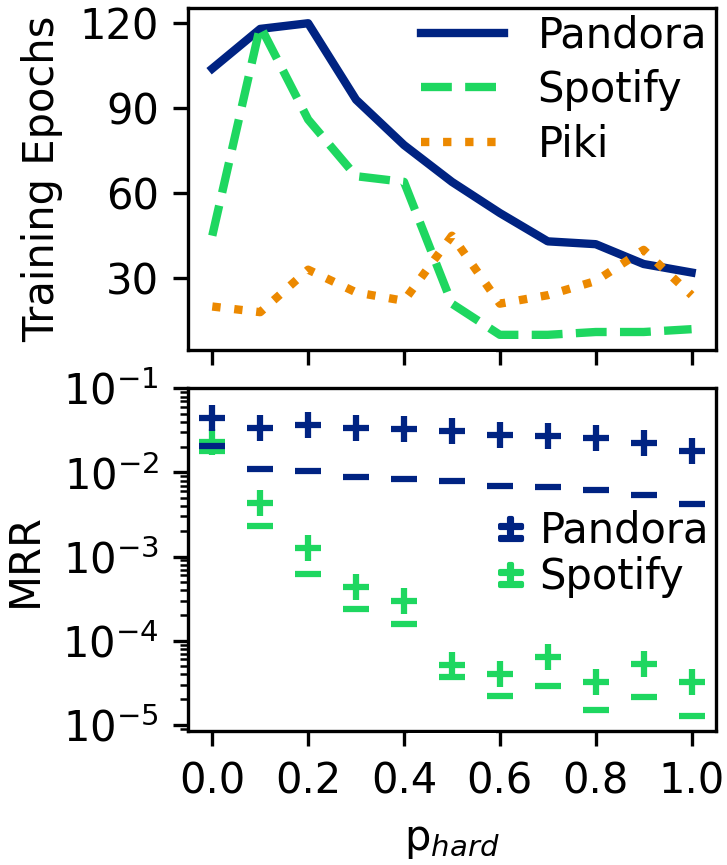}
  \caption{For higher $p_{hard}$, training time tends to decrease (top), while the mean reciprocal ranks of both the positive ($+$) and negative ($-$) samples generally decrease (bottom), even as the test accuracy increases (Fig. \ref{fig:hard_negs}). 
   The Piki dataset is omitted as it is too small/noisy.
  }
  \Description{Top: A chart showing how training time decreases when using higher $p_{hard}$. Bottom: A chart showing how the ranks of both the positive and negative songs decline as $p_{hard}$ increases.}
  \label{fig:epochs}
\end{wrapfigure}
\fi 
The test accuracy is considerably higher, though quickly plateaus, for any $p_{hard} \neq 0$ (Fig. \ref{fig:hard_negs}). 
This is likely because at each epoch, the specific $p_{task}$ positions that see hard negatives are re-sampled with probability $p_{hard}$ and so even with lower $p_{hard}$ values, eventually all positions that have hard negatives will use them for some part of the training time. 
However, higher $p_{hard}$ values allow the model to use more of these per epoch, and hence the model can converge faster.

\subsubsection{Training time}

We find that the number of epochs necessary for convergence reduces considerably for higher values of $p_{hard}$ (Fig. \ref{fig:epochs}). This is less noticeable/relevant for the Piki dataset because it is so small. 
Weighing this against the plateau in accuracy for higher $p_{hard}$, this suggests that higher values of $p_{hard}$ are better. 
However, this may lead to poorer embeddings and hence poorer recommendation quality for more tail-end songs, which are more rarely encountered (as random negative samples) in the training set.

% \ifdim\columnwidth<0.7\textwidth
% \begin{figure}[b]
%     \includegraphics[width=0.35\textwidth]{figs/mrr.png}
%   \caption{The mean reciprocal rank of the positive and hard negative samples generally both decrease as the $p_{hard}$ used during training increases. The prediction accuracy is somewhat connected to the gap between the ranks of the positive and negative samples.}
%   % \Description{A chart showing the mean reciprocal rank for models trained with different values of $p_{hard}$. As $p_{hard}$ increases, the mean reciprocal rank decreases for both the up and down thumbs, with the up thumbs decreasing faster than down thumbs.}
%   \label{fig:mrr}
% \end{figure}
% \else
% \begin{wrapfigure}{r}{0.31\textwidth}
%   \centering
%     \includegraphics[width=0.3\textwidth]{figs/mrr_combined_small.png}
%   \caption{The mean reciprocal ranks of the positive ($+$) and negative ($-$) samples generally both decrease, even as the test accuracy increases (Fig. \ref{fig:hard_negs}), as the $p_{hard}$ used during training increases. 
%   % The prediction accuracy is somewhat connected to the gap between the ranks of the positive and negative samples.
%   }
%     % \Description{A chart showing the mean reciprocal rank for models trained with different values of $p_{hard}$. As $p_{hard}$ increases, the mean reciprocal rank decreases for both the up and down thumbs, with the up thumbs decreasing faster than down thumbs.}
%   \label{fig:mrr}
% \end{wrapfigure}
% \fi
\subsubsection{Relation to other metrics}\label{discuss:mrr}
The prediction task of comparing a user's positive and negative feedback is generally much harder than comparing a user's positive feedback to a random negative sample. 
The latter is somewhat connected to true positive metrics such as recall or the mean reciprocal rank (MRR). 
However, in our use case, the songs eligible to be ranked on a given station are generally similarly-popular songs, whether positive or negative.

The MRR (and the recall) of both the `up' and `down' samples decrease with higher $p_{hard}$, though the `down' samples decrease faster than the `up' samples (Fig. \ref{fig:epochs}), such that the overall test accuracy increases (Fig. \ref{fig:hard_negs}). 
The prediction accuracy is somewhat connected to the distance between the positive and negative ranks. 
This can be interpreted as higher $p_{hard}$ attempting to minimize false positives (at the cost of fewer true positives), or inversely, that lower $p_{hard}$ maximizes true positives but also increases false positives. 
Similar to Mena-Maldonado et al. (2020), we find that reducing false positives with a higher $p_{hard}$ also leads to generally less-popular songs being recommended \cite{mena2020}. 

For candidate generation, it may be better to use a low $p_{hard}$ model which pushes all potentially likeable songs to the top of the rankings (even if some are false positives). 
For ranking some pre-existing song candidates (which could be generated by a low $p_{hard}$ model), it may be better to use a higher $p_{hard}$ model that pushes down bad songs for the user.

\subsection{Feedback types}\label{discuss:ftype}
\ifdim\columnwidth<0.7\textwidth
\begin{figure}[btp]
  \includegraphics[width=0.33\textwidth]{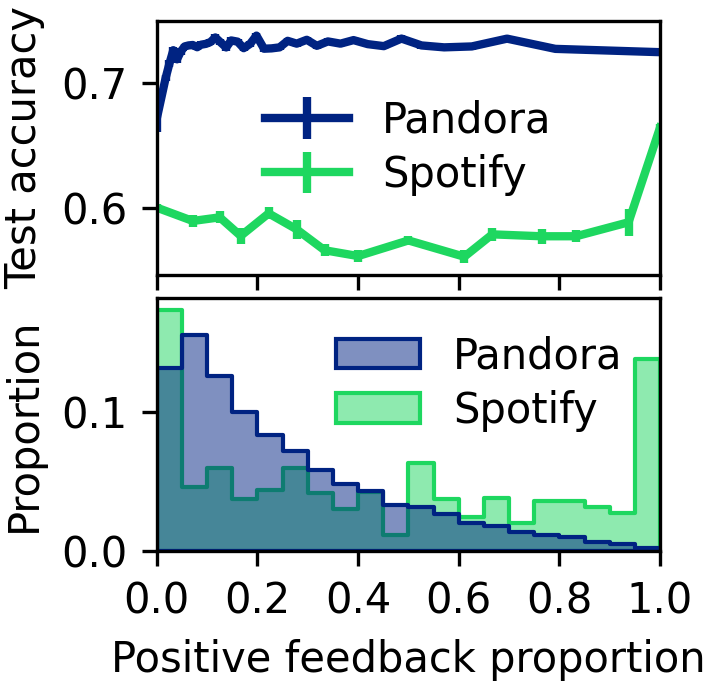}
  \caption{Test accuracy is fairly consistent across users with different proportion of feedback types (top). The distribution of feedback proportions differs by dataset (bottom).}
  % \Description{(a) a chart showing the distribution of input lengths. It is skewed towards shorter lengths and has an artificial spike at x=400 due to truncation of the inputs. (b) A chart showing the relative ratios of skips and up thumbs at each position; earlier positions have more skips. (c) A chart showing the distribution of skip proportions for different input lengths. Users with more inputs are more likely to have more skips.}
  \label{fig:skipratio}
\end{figure}
\else
\begin{wrapfigure}[15]{r}{0.335\textwidth}
  \centering
  \includegraphics[width=0.32\textwidth]{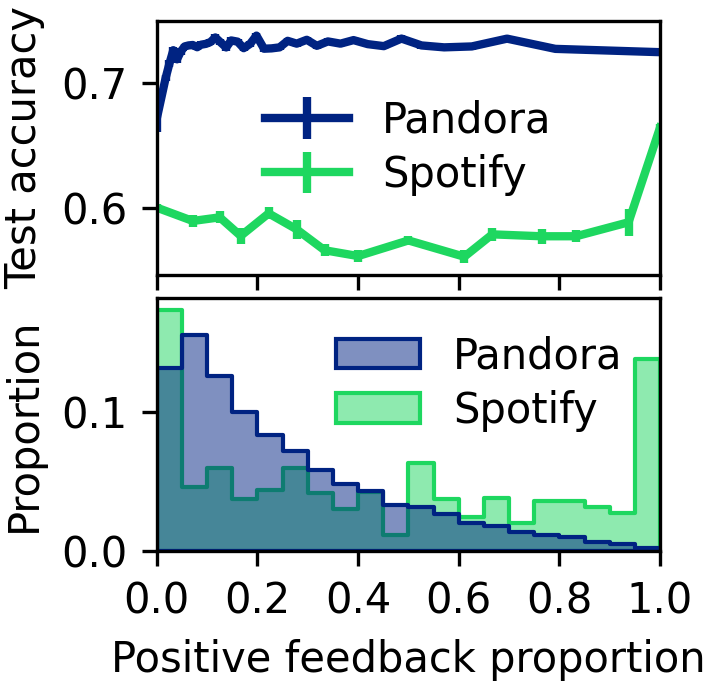}
  \caption{Test accuracy is fairly consistent across users with different proportion of feedback types (top). The distribution of feedback proportions differs by dataset (bottom).}
  % \Description{(a) a chart showing the distribution of input lengths. It is skewed towards shorter lengths and has an artificial spike at x=400 due to truncation of the inputs. (b) A chart showing the relative ratios of skips and up thumbs at each position; earlier positions have more skips. (c) A chart showing the distribution of skip proportions for different input lengths. Users with more inputs are more likely to have more skips.}
  \label{fig:skipratio}
\end{wrapfigure}
\fi

% \ifdim\columnwidth<0.7\textwidth
% \begin{figure}[h!]
%   \includegraphics[width=0.45\textwidth]{figs/feed_type_embs.png}
%   \caption{Top: The raw embedding weights for the different feedback types for each hidden dimension. Bottom: A comparison of the `skip' embedding vs. the `up' and `down' embeddings for each embedding dimension, with the correlation coefficient ($R$) also shown.}
%   % \Description{A }
%   \label{fig:feed_types}
% \end{figure}
% \else
% \begin{wrapfigure}{r}{0.4\textwidth}
%   \centering
%     \includegraphics[width=0.4\textwidth]{figs/feed_type_embs.png}
%   \caption{Top: The raw embedding weights for the different feedback types for each hidden dimension. Bottom: A comparison of the `skip' embedding vs. the `up' and `down' embeddings for each embedding dimension, with the correlation coefficient ($R$) also shown.}
%   % \Description{A }
%   \label{fig:feed_types}
% \end{wrapfigure}
% \fi

The Pandora dataset includes three different types of feedback and so can be used to identify some qualitative aspects of `skip' feedback. 
We have generally asserted that skips are implicit negative feedback, but it is also possible that they share some aspects of positive feedback, as users may thumb up a song, then skip it if it is replayed too much. 
The raw embeddings for `skip' feedback are more similar to `down' feedback (cosine similarity: $0.82 \pm 0.01$ when averaged over all $p_{hard} > 0$ models)
% (Fig. \ref{fig:feed_types}),
than to `up' feedback (similarity $0.16 \pm 0.01$), although the fact that the model does not learn identical embeddings for `skips' and `downs' suggests that they are still distinct in some way. 
The  similarity for the `up' and `down' embeddings is $0.00 \pm 0.02$. The Spotify `up' (i.e. `play') and `skip' embeddings have an average similarity of $-0.53 \pm 0.08$; the Piki `up' and `down' embeddings have an average similarity of $0.39 \pm 0.03$. 
This may suggest that users may use feedback features differently between these datasets.

The test accuracy is generally consistent across users with different proportions of positive feedback (Fig. \ref{fig:skipratio}). Pandora users with almost no positive feedback (i.e. `mostly-skip' users) have a slightly lower test accuracy by $\sim2\%$; inversely, Spotify users with almost no skips (i.e. `mostly-play' users) have a higher test accuracy by $\sim6\%$. 
Further work should be done to better understand these user segments; we note that the bimodal distribution of positive feedback proportions in the Spotify dataset (Fig. \ref{fig:skipratio}) aligns with prior research clustering user types into `mostly-play' and `mostly-skip' \cite{meggetto2021}.

\section{Conclusion}
We find that incorporating user-given hard negatives in training considerably improves the test accuracy of user-liked songs as compared to using random negative samples.
An approximation to a hard negative can be found by sampling multiple negatives and choosing the hardest one: we find that this can help test accuracy up to a point, but too many random negatives are likely to include some false negatives which reduces the test accuracy. 
As many users give `skip' feedback, this can be used directly in the input to help define a user's context vector, which slightly improves the overall accuracy of the model and also considerably improves the user coverage compared to only using positive feedback.
We also find that the test accuracy is robust with respect to differing proportions of feedback types, and that `skip' feedback is more similar to negative feedback than positive feedback.
% Lastly, we find that models trained with hard negatives decay in accuracy more slowly than those trained with random negative samples.
% \vspace{-3mm}

% \begin{acks}
% \end{acks}
\bibliographystyle{ACM-Reference-Format}
\bibliography{stamper-bib}


\begin{thebibliography}{23}

%%% the Web address.

\ifx \showCODEN    \undefined \def \showCODEN     #1{\unskip}     \fi
\ifx \showDOI      \undefined \def \showDOI       #1{#1}\fi
\ifx \showISBNx    \undefined \def \showISBNx     #1{\unskip}     \fi
\ifx \showISBNxiii \undefined \def \showISBNxiii  #1{\unskip}     \fi
\ifx \showISSN     \undefined \def \showISSN      #1{\unskip}     \fi
\ifx \showLCCN     \undefined \def \showLCCN      #1{\unskip}     \fi
\ifx \shownote     \undefined \def \shownote      #1{#1}          \fi
\ifx \showarticletitle \undefined \def \showarticletitle #1{#1}   \fi
\ifx \showURL      \undefined \def \showURL       {\relax}        \fi
% invisible to TeX
\providecommand\bibfield[2]{#2}
\providecommand\bibinfo[2]{#2}
\providecommand\natexlab[1]{#1}
\providecommand\showeprint[2][]{arXiv:#2}

\bibitem[Bengio and Senecal(2008)]%
        {bengio2008}
\bibfield{author}{\bibinfo{person}{Yoshua Bengio} {and} \bibinfo{person}{Jean-SÉbastien Senecal}.} \bibinfo{year}{2008}\natexlab{}.
\newblock \showarticletitle{Adaptive Importance Sampling to Accelerate Training of a Neural Probabilistic Language Model}.
\newblock \bibinfo{journal}{\emph{IEEE Transactions on Neural Networks}} \bibinfo{volume}{19}, \bibinfo{number}{4} (\bibinfo{year}{2008}), \bibinfo{pages}{713--722}.
\newblock
\urldef\tempurl%
\url{https://doi.org/10.1109/TNN.2007.912312}
\showDOI{\tempurl}


\bibitem[Brost et~al\mbox{.}(2019)]%
        {brost2019}
\bibfield{author}{\bibinfo{person}{Brian Brost}, \bibinfo{person}{Rishabh Mehrotra}, {and} \bibinfo{person}{Tristan Jehan}.} \bibinfo{year}{2019}\natexlab{}.
\newblock \showarticletitle{The Music Streaming Sessions Dataset}. In \bibinfo{booktitle}{\emph{Proceedings of the 2019 Web Conference}}. ACM.
\newblock


\bibitem[Dias and Fonseca(2013)]%
        {dias2013}
\bibfield{author}{\bibinfo{person}{Ricardo Dias} {and} \bibinfo{person}{Manuel~J Fonseca}.} \bibinfo{year}{2013}\natexlab{}.
\newblock \showarticletitle{Improving music recommendation in session-based collaborative filtering by using temporal context}. In \bibinfo{booktitle}{\emph{2013 IEEE 25th international conference on tools with artificial intelligence}}. IEEE, \bibinfo{pages}{783--788}.
\newblock
\urldef\tempurl%
\url{https://doi.org/10.1109/ICTAI.2013.120}
\showDOI{\tempurl}


\bibitem[Hanley and McNeil(1982)]%
        {hanley1982}
\bibfield{author}{\bibinfo{person}{James~A Hanley} {and} \bibinfo{person}{Barbara~J McNeil}.} \bibinfo{year}{1982}\natexlab{}.
\newblock \showarticletitle{The meaning and use of the area under a receiver operating characteristic (ROC) curve.}
\newblock \bibinfo{journal}{\emph{Radiology}} \bibinfo{volume}{143}, \bibinfo{number}{1} (\bibinfo{year}{1982}), \bibinfo{pages}{29--36}.
\newblock


\bibitem[Kang and McAuley(2018)]%
        {kang2018}
\bibfield{author}{\bibinfo{person}{Wang-Cheng Kang} {and} \bibinfo{person}{Julian McAuley}.} \bibinfo{year}{2018}\natexlab{}.
\newblock \showarticletitle{Self-attentive sequential recommendation}. In \bibinfo{booktitle}{\emph{2018 IEEE International Conference on Data Mining (ICDM)}}. IEEE, \bibinfo{pages}{197--206}.
\newblock
\urldef\tempurl%
\url{https://doi.org/10.1109/ICDM.2018.00035}
\showDOI{\tempurl}


\bibitem[Meggetto et~al\mbox{.}(2021)]%
        {meggetto2021}
\bibfield{author}{\bibinfo{person}{Francesco Meggetto}, \bibinfo{person}{Crawford Revie}, \bibinfo{person}{John Levine}, {and} \bibinfo{person}{Yashar Moshfeghi}.} \bibinfo{year}{2021}\natexlab{}.
\newblock \showarticletitle{On Skipping Behaviour Types in Music Streaming Sessions}. In \bibinfo{booktitle}{\emph{Proceedings of the 30th ACM International Conference on Information \& Knowledge Management}} (Virtual Event, Queensland, Australia) \emph{(\bibinfo{series}{CIKM '21})}. \bibinfo{publisher}{Association for Computing Machinery}, \bibinfo{address}{New York, NY, USA}, \bibinfo{pages}{3333–3337}.
\newblock
\showISBNx{9781450384469}
\urldef\tempurl%
\url{https://doi.org/10.1145/3459637.3482123}
\showDOI{\tempurl}


\bibitem[Meggetto et~al\mbox{.}(2023)]%
        {meggetto2023}
\bibfield{author}{\bibinfo{person}{Francesco Meggetto}, \bibinfo{person}{Crawford Revie}, \bibinfo{person}{John Levine}, {and} \bibinfo{person}{Yashar Moshfeghi}.} \bibinfo{year}{2023}\natexlab{}.
\newblock \showarticletitle{Why People Skip Music? On Predicting Music Skips using Deep Reinforcement Learning}. In \bibinfo{booktitle}{\emph{Proceedings of the 2023 Conference on Human Information Interaction and Retrieval}} (<conf-loc>, <city>Austin</city>, <state>TX</state>, <country>USA</country>, </conf-loc>) \emph{(\bibinfo{series}{CHIIR '23})}. \bibinfo{publisher}{Association for Computing Machinery}, \bibinfo{address}{New York, NY, USA}, \bibinfo{pages}{95–106}.
\newblock
\showISBNx{9798400700354}
\urldef\tempurl%
\url{https://doi.org/10.1145/3576840.3578312}
\showDOI{\tempurl}


\bibitem[Mei et~al\mbox{.}(2023)]%
        {mei2023}
\bibfield{author}{\bibinfo{person}{M.~Jeffrey Mei}, \bibinfo{person}{Oliver Bembom}, {and} \bibinfo{person}{Andreas Ehmann}.} \bibinfo{year}{2023}\natexlab{}.
\newblock \showarticletitle{Station and Track Attribute-Aware Music Personalization}. In \bibinfo{booktitle}{\emph{Proceedings of the 17th ACM Conference on Recommender Systems}} (Singapore, Singapore) \emph{(\bibinfo{series}{RecSys '23})}. \bibinfo{publisher}{Association for Computing Machinery}, \bibinfo{address}{New York, NY, USA}, \bibinfo{pages}{1031–1035}.
\newblock
\showISBNx{9798400702419}
\urldef\tempurl%
\url{https://doi.org/10.1145/3604915.3610239}
\showDOI{\tempurl}


\bibitem[Mena-Maldonado et~al\mbox{.}(2020)]%
        {mena2020}
\bibfield{author}{\bibinfo{person}{Elisa Mena-Maldonado}, \bibinfo{person}{Roc\'{\i}o Ca\~{n}amares}, \bibinfo{person}{Pablo Castells}, \bibinfo{person}{Yongli Ren}, {and} \bibinfo{person}{Mark Sanderson}.} \bibinfo{year}{2020}\natexlab{}.
\newblock \showarticletitle{Agreement and Disagreement between True and False-Positive Metrics in Recommender Systems Evaluation}. In \bibinfo{booktitle}{\emph{Proceedings of the 43rd International ACM SIGIR Conference on Research and Development in Information Retrieval}} (Virtual Event, China) \emph{(\bibinfo{series}{SIGIR '20})}. \bibinfo{publisher}{Association for Computing Machinery}, \bibinfo{address}{New York, NY, USA}, \bibinfo{pages}{841–850}.
\newblock
\showISBNx{9781450380164}
\urldef\tempurl%
\url{https://doi.org/10.1145/3397271.3401096}
\showDOI{\tempurl}


\bibitem[Petrov and Macdonald(2023)]%
        {petrov2023}
\bibfield{author}{\bibinfo{person}{Aleksandr~Vladimirovich Petrov} {and} \bibinfo{person}{Craig Macdonald}.} \bibinfo{year}{2023}\natexlab{}.
\newblock \showarticletitle{GSASRec: Reducing Overconfidence in Sequential Recommendation Trained with Negative Sampling}. In \bibinfo{booktitle}{\emph{Proceedings of the 17th ACM Conference on Recommender Systems}} (Singapore, Singapore) \emph{(\bibinfo{series}{RecSys '23})}. \bibinfo{publisher}{Association for Computing Machinery}, \bibinfo{address}{New York, NY, USA}, \bibinfo{pages}{116–128}.
\newblock
\showISBNx{9798400702419}
\urldef\tempurl%
\url{https://doi.org/10.1145/3604915.3608783}
\showDOI{\tempurl}


\bibitem[Schedl and Schnitzer(2014)]%
        {schedl2014}
\bibfield{author}{\bibinfo{person}{Markus Schedl} {and} \bibinfo{person}{Dominik Schnitzer}.} \bibinfo{year}{2014}\natexlab{}.
\newblock \showarticletitle{Location-aware music artist recommendation}. In \bibinfo{booktitle}{\emph{MultiMedia Modeling: 20th Anniversary International Conference, MMM 2014, Dublin, Ireland, January 6-10, 2014, Proceedings, Part II 20}}. Springer, \bibinfo{pages}{205--213}.
\newblock


\bibitem[Seshadri and Knees(2023)]%
        {seshadri2023}
\bibfield{author}{\bibinfo{person}{Pavan Seshadri} {and} \bibinfo{person}{Peter Knees}.} \bibinfo{year}{2023}\natexlab{}.
\newblock \showarticletitle{Leveraging Negative Signals with Self-Attention for Sequential Music Recommendation}.
\newblock \bibinfo{journal}{\emph{arXiv preprint arXiv:2309.11623}} (\bibinfo{year}{2023}).
\newblock


\bibitem[Stoikov and Wen(2021)]%
        {stoikov2021}
\bibfield{author}{\bibinfo{person}{Sasha Stoikov} {and} \bibinfo{person}{Hongyi Wen}.} \bibinfo{year}{2021}\natexlab{}.
\newblock \showarticletitle{Evaluating Music Recommendations with Binary Feedback for Multiple Stakeholders}. In \bibinfo{booktitle}{\emph{Proceedings of the 1st Workshop on Multi-Objective Recommender Systems (MORS 2021) co-located with 15th ACM Conference on Recommender Systems (RecSys 2021)}} \emph{(\bibinfo{series}{{CEUR} Workshop Proceedings}, Vol.~\bibinfo{volume}{2959})}. \bibinfo{publisher}{CEUR-WS.org}.
\newblock
\urldef\tempurl%
\url{https://ceur-ws.org/Vol-2959/paper9.pdf}
\showURL{%
\tempurl}


\bibitem[Sun et~al\mbox{.}(2019)]%
        {sun2019}
\bibfield{author}{\bibinfo{person}{Fei Sun}, \bibinfo{person}{Jun Liu}, \bibinfo{person}{Jian Wu}, \bibinfo{person}{Changhua Pei}, \bibinfo{person}{Xiao Lin}, \bibinfo{person}{Wenwu Ou}, {and} \bibinfo{person}{Peng Jiang}.} \bibinfo{year}{2019}\natexlab{}.
\newblock \showarticletitle{BERT4Rec: Sequential Recommendation with Bidirectional Encoder Representations from Transformer}. In \bibinfo{booktitle}{\emph{Proceedings of the 28th ACM International Conference on Information and Knowledge Management}} (Beijing, China) \emph{(\bibinfo{series}{CIKM '19})}. \bibinfo{publisher}{Association for Computing Machinery}, \bibinfo{address}{New York, NY, USA}, \bibinfo{pages}{1441–1450}.
\newblock
\showISBNx{9781450369763}
\urldef\tempurl%
\url{https://doi.org/10.1145/3357384.3357895}
\showDOI{\tempurl}


\bibitem[Vaswani et~al\mbox{.}(2017)]%
        {vaswani2017}
\bibfield{author}{\bibinfo{person}{Ashish Vaswani}, \bibinfo{person}{Noam Shazeer}, \bibinfo{person}{Niki Parmar}, \bibinfo{person}{Jakob Uszkoreit}, \bibinfo{person}{Llion Jones}, \bibinfo{person}{Aidan~N Gomez}, \bibinfo{person}{{\L}ukasz Kaiser}, {and} \bibinfo{person}{Illia Polosukhin}.} \bibinfo{year}{2017}\natexlab{}.
\newblock \showarticletitle{Attention is all you need}.
\newblock \bibinfo{journal}{\emph{Advances in Neural Information Processing Systems}}  \bibinfo{volume}{30} (\bibinfo{year}{2017}).
\newblock
\urldef\tempurl%
\url{https://doi.org/10.48550/arXiv.1706.03762}
\showDOI{\tempurl}


\bibitem[Wang et~al\mbox{.}(2018)]%
        {wang2018}
\bibfield{author}{\bibinfo{person}{Jizhe Wang}, \bibinfo{person}{Pipei Huang}, \bibinfo{person}{Huan Zhao}, \bibinfo{person}{Zhibo Zhang}, \bibinfo{person}{Binqiang Zhao}, {and} \bibinfo{person}{Dik~Lun Lee}.} \bibinfo{year}{2018}\natexlab{}.
\newblock \showarticletitle{Billion-Scale Commodity Embedding for E-Commerce Recommendation in Alibaba}. In \bibinfo{booktitle}{\emph{Proceedings of the 24th ACM SIGKDD International Conference on Knowledge Discovery \& Data Mining}} (London, United Kingdom) \emph{(\bibinfo{series}{KDD '18})}. \bibinfo{publisher}{Association for Computing Machinery}, \bibinfo{address}{New York, NY, USA}, \bibinfo{pages}{839–848}.
\newblock
\showISBNx{9781450355520}
\urldef\tempurl%
\url{https://doi.org/10.1145/3219819.3219869}
\showDOI{\tempurl}


\bibitem[Wang and Wang(2014)]%
        {wang2014}
\bibfield{author}{\bibinfo{person}{Xinxi Wang} {and} \bibinfo{person}{Ye Wang}.} \bibinfo{year}{2014}\natexlab{}.
\newblock \showarticletitle{Improving content-based and hybrid music recommendation using deep learning}. In \bibinfo{booktitle}{\emph{Proceedings of the 22nd ACM international conference on Multimedia}}. \bibinfo{pages}{627--636}.
\newblock
\urldef\tempurl%
\url{https://doi.org/10.1145/2647868.2654940}
\showDOI{\tempurl}


\bibitem[Wang et~al\mbox{.}(2023)]%
        {wang2023}
\bibfield{author}{\bibinfo{person}{Yueqi Wang}, \bibinfo{person}{Yoni Halpern}, \bibinfo{person}{Shuo Chang}, \bibinfo{person}{Jingchen Feng}, \bibinfo{person}{Elaine~Ya Le}, \bibinfo{person}{Longfei Li}, \bibinfo{person}{Xujian Liang}, \bibinfo{person}{Min-Cheng Huang}, \bibinfo{person}{Shane Li}, \bibinfo{person}{Alex Beutel}, \bibinfo{person}{Yaping Zhang}, {and} \bibinfo{person}{Shuchao Bi}.} \bibinfo{year}{2023}\natexlab{}.
\newblock \showarticletitle{Learning from Negative User Feedback and Measuring Responsiveness for Sequential Recommenders}. In \bibinfo{booktitle}{\emph{Proceedings of the 17th ACM Conference on Recommender Systems}} (Singapore, Singapore) \emph{(\bibinfo{series}{RecSys '23})}. \bibinfo{publisher}{Association for Computing Machinery}, \bibinfo{address}{New York, NY, USA}, \bibinfo{pages}{1049–1053}.
\newblock
\showISBNx{9798400702419}
\urldef\tempurl%
\url{https://doi.org/10.1145/3604915.3610244}
\showDOI{\tempurl}


\bibitem[Wen et~al\mbox{.}(2019)]%
        {wen2019}
\bibfield{author}{\bibinfo{person}{Hongyi Wen}, \bibinfo{person}{Longqi Yang}, {and} \bibinfo{person}{Deborah Estrin}.} \bibinfo{year}{2019}\natexlab{}.
\newblock \showarticletitle{Leveraging post-click feedback for content recommendations}. In \bibinfo{booktitle}{\emph{Proceedings of the 13th ACM Conference on Recommender Systems}} (Copenhagen, Denmark) \emph{(\bibinfo{series}{RecSys '19})}. \bibinfo{publisher}{Association for Computing Machinery}, \bibinfo{address}{New York, NY, USA}, \bibinfo{pages}{278–286}.
\newblock
\showISBNx{9781450362436}
\urldef\tempurl%
\url{https://doi.org/10.1145/3298689.3347037}
\showDOI{\tempurl}


\bibitem[Weston et~al\mbox{.}(2011)]%
        {weston2011}
\bibfield{author}{\bibinfo{person}{Jason Weston}, \bibinfo{person}{Samy Bengio}, {and} \bibinfo{person}{Nicolas Usunier}.} \bibinfo{year}{2011}\natexlab{}.
\newblock \showarticletitle{Wsabie: Scaling Up To Large Vocabulary Image Annotation}. In \bibinfo{booktitle}{\emph{Proceedings of the International Joint Conference on Artificial Intelligence, IJCAI}}.
\newblock


\bibitem[Wilm et~al\mbox{.}(2023)]%
        {wilm2023}
\bibfield{author}{\bibinfo{person}{Timo Wilm}, \bibinfo{person}{Philipp Normann}, \bibinfo{person}{Sophie Baumeister}, {and} \bibinfo{person}{Paul-Vincent Kobow}.} \bibinfo{year}{2023}\natexlab{}.
\newblock \showarticletitle{Scaling Session-Based Transformer Recommendations Using Optimized Negative Sampling and Loss Functions} \emph{(\bibinfo{series}{RecSys '23})}. \bibinfo{publisher}{Association for Computing Machinery}, \bibinfo{address}{New York, NY, USA}, \bibinfo{pages}{1023–1026}.
\newblock
\showISBNx{9798400702419}
\urldef\tempurl%
\url{https://doi.org/10.1145/3604915.3610236}
\showDOI{\tempurl}


\bibitem[Wu et~al\mbox{.}(2023)]%
        {wu2023}
\bibfield{author}{\bibinfo{person}{Jiancan Wu}, \bibinfo{person}{Xiang Wang}, \bibinfo{person}{Xingyu Gao}, \bibinfo{person}{Jiawei Chen}, \bibinfo{person}{Hongcheng Fu}, \bibinfo{person}{Tianyu Qiu}, {and} \bibinfo{person}{Xiangnan He}.} \bibinfo{year}{2023}\natexlab{}.
\newblock \showarticletitle{On the Effectiveness of Sampled Softmax Loss for Item Recommendation}.
\newblock \bibinfo{journal}{\emph{ACM Trans. Inf. Syst.}} (\bibinfo{date}{dec} \bibinfo{year}{2023}).
\newblock
\showISSN{1046-8188}
\urldef\tempurl%
\url{https://doi.org/10.1145/3637061}
\showDOI{\tempurl}
\newblock
\shownote{Just Accepted}.


\bibitem[Zhao et~al\mbox{.}(2023)]%
        {zhao2023}
\bibfield{author}{\bibinfo{person}{Yuhan Zhao}, \bibinfo{person}{Rui Chen}, \bibinfo{person}{Riwei Lai}, \bibinfo{person}{Qilong Han}, \bibinfo{person}{Hongtao Song}, {and} \bibinfo{person}{Li Chen}.} \bibinfo{year}{2023}\natexlab{}.
\newblock \showarticletitle{Augmented Negative Sampling for Collaborative Filtering}. In \bibinfo{booktitle}{\emph{Proceedings of the 17th ACM Conference on Recommender Systems}} (Singapore, Singapore) \emph{(\bibinfo{series}{RecSys '23})}. \bibinfo{publisher}{Association for Computing Machinery}, \bibinfo{address}{New York, NY, USA}, \bibinfo{pages}{256–266}.
\newblock
\showISBNx{9798400702419}
\urldef\tempurl%
\url{https://doi.org/10.1145/3604915.3608811}
\showDOI{\tempurl}


\end{thebibliography}

% \appendix









\end{document}